\documentclass[10pt,twocolumn,letterpaper]{article}

\usepackage{cvpr}              % To produce the CAMERA-READY version
\usepackage{graphicx}
\usepackage{amsmath}
\usepackage{amssymb}
\usepackage{booktabs}
\usepackage{wrapfig}
\usepackage{cite}
\usepackage{multirow}
\usepackage{makecell}
\usepackage{booktabs}       % professional-quality tables
\usepackage{xcolor}         % colors
\usepackage{bm}
\usepackage{float}
\usepackage{mathtools}
\usepackage{caption}
\usepackage{paralist}
\usepackage[accsupp]{axessibility}  % Improves PDF readability for those with disabilities.

\usepackage[pagebackref,breaklinks,colorlinks]{hyperref}
\usepackage[normalem]{ulem}

\usepackage[capitalize]{cleveref}
\crefname{section}{Sec.}{Secs.}
\Crefname{section}{Section}{Sections}
\Crefname{table}{Table}{Tables}
\crefname{table}{Tab.}{Tabs.}

\def\cvprPaperID{2630} % *** Enter the CVPR Paper ID here
\def\confName{CVPR}
\def\confYear{2023}

\begin{document}

% Skinned Motion Retargeting with Residual Perception of Motion Semantics \& Geometry
% Differentiable Motion Semantics \& Geometry Perception \\ for Residual Skinned Motion Retargeting
\title{Skinned Motion Retargeting with \\ Residual Perception of Motion Semantics \& Geometry}

% \and
% } \thanks{Corresponding author: tuzhigang@whu.edu.cn}\qquad \thanks{Work done at Wuhan University}\qquad

\author{
    Jiaxu Zhang$^{1}$\thanks{Most of this work was done during Jiaxu’s internship at Tencent.} \qquad
    Junwu Weng$^{2}$ \qquad
    Di Kang$^{2}$\qquad
    Fang Zhao$^{2}$\qquad
    Shaoli Huang$^{2}$\qquad \\
    Xuefei Zhe$^{2}$\qquad
    Linchao Bao$^{2}$\qquad
    Ying Shan$^{2}$\qquad
    Jue Wang$^{2}$\qquad
    Zhigang Tu$^{1}$\thanks{Corresponding author: tuzhigang@whu.edu.cn}\\ 
 $^{1}$Wuhan University\qquad
 $^{2}$Tencent AI Lab \qquad
    {\tt\small \{zjiaxu, tuzhigang\}@whu.edu.cn}
    }

%   \includegraphics[width=\textwidth]{figures/figure1.pdf}

% \twocolumn[{%
% \renewcommand\twocolumn[1][]{#1}%
% \maketitle
% \begin{figure}[H]
% \hsize=\textwidth
% \vspace{-1.3cm}
% \setlength{\abovecaptionskip}{-.2cm}
% % \setlength{\belowcaptionskip}{-.2cm}
% \begin{center}
%    \includegraphics[width=0.95\textwidth]{figures/figure1.pdf}
% \end{center}
%    \caption{Previous methods, e.g. SAN \cite{aberman2020skeleton}, are prone to produce unrealistic results when retargeting motions on two characters with different kinematic configurations (left) and geometries (right). Our CRet preserves motion semantics, eliminates self-penetration, and keeps self-contact without post-optimizations. The retargeted motion of our CRet is continuously controllable for animators to edit interactively.}
% \label{fig:1}
% \end{figure}
% }]

\maketitle

%%%%%%%%% ABSTRACT
\begin{abstract}
% \vspace{-2mm}
A good motion retargeting cannot be reached without reasonable consideration of source-target differences on both the skeleton and shape geometry levels. In this work, we propose a novel \textbf{R}esidual \textbf{RET}argeting network (R$^2$ET) structure, which relies on two neural modification modules, to adjust the source motions to fit the target skeletons and shapes progressively. In particular, a skeleton-aware module is introduced to preserve the source motion semantics. A shape-aware module is designed to perceive the geometries of target characters to reduce interpenetration and contact-missing. Driven by our explored distance-based losses that explicitly model the motion semantics and geometry, these two modules can learn residual motion modifications on the source motion to generate plausible retargeted motion in a single inference without post-processing. To balance these two modifications, we further present a balancing gate to conduct linear interpolation between them. Extensive experiments on the public dataset Mixamo demonstrate that our R$^2$ET achieves the state-of-the-art performance, and provides a good balance between the preservation of motion semantics as well as the attenuation of interpenetration and contact-missing. Code is available at \url{https://github.com/Kebii/R2ET}.
\end{abstract}

%%%%%%%%% BODY TEXT
\vspace{-2.5mm}
\section{Introduction}
\vspace{-1.0mm}

\begin{figure}[t]
\vspace{-.2cm}
\setlength{\abovecaptionskip}{-.1cm}
\setlength{\belowcaptionskip}{-.3cm}
\begin{center}
   \includegraphics[width=0.95\linewidth]{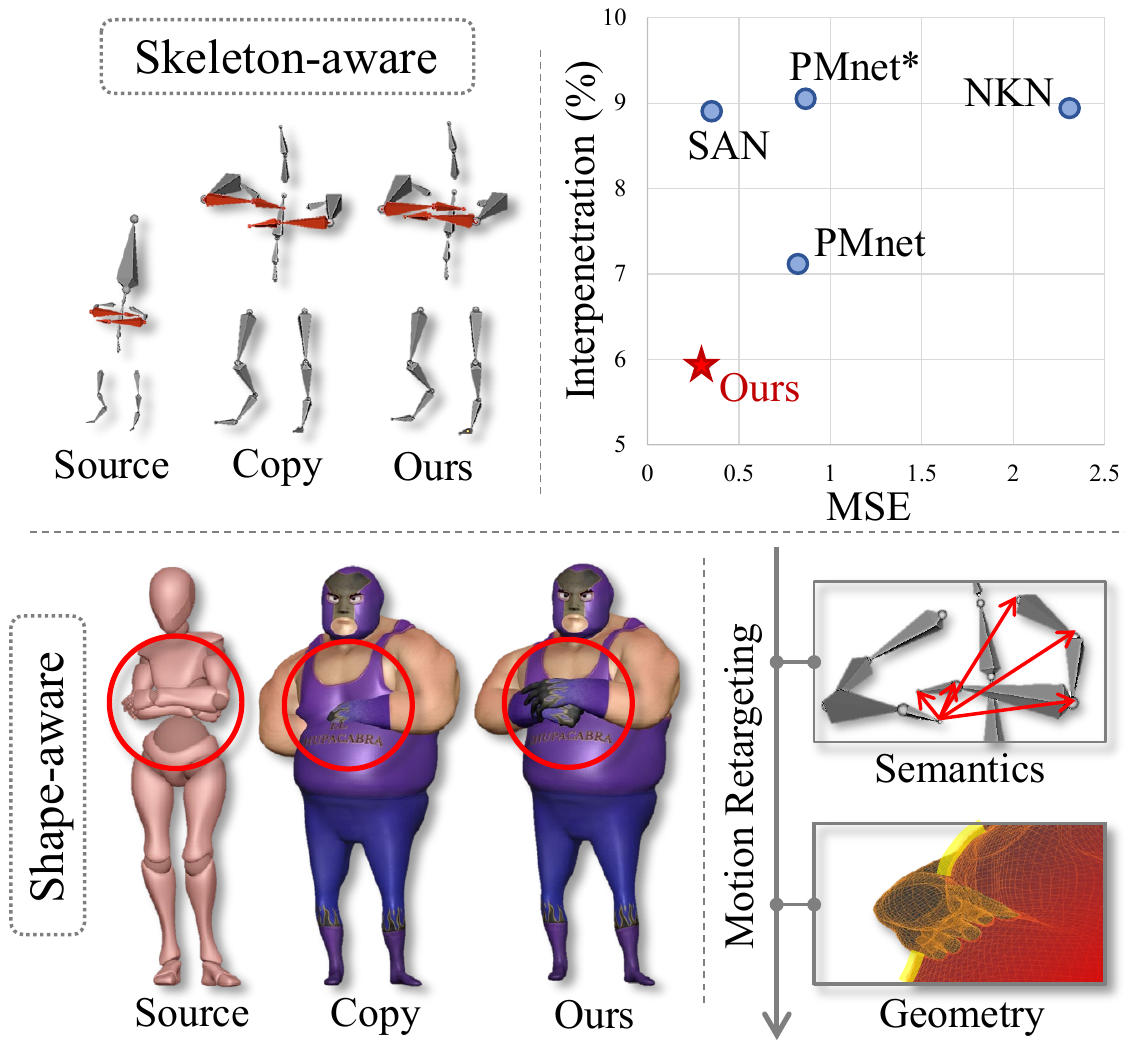}
\end{center}
   \caption{Our R$^2$ET fully considers the source-target differences on both the skeleton and shape geometry levels. The retargeted motion of R$^2$ET preserves motion semantics, eliminates interpenetration, and keeps self-contact without post-optimizations.}
\label{fig:1}
\end{figure}

As a process of mapping the motion of a source character to a target character without losing plausibility, \textit{motion retargeting} is a long-standing problem in the community of computer vision and computer graphics. It has a wide spectrum of applications in game and animation industry, and is a cornerstone of the digital avatar and metaverse technologies~\cite{villegas2018neural}. In recent years, learning-based retargeting methods started sparkling in the community. Among them, the neural motion retargeting~\cite{villegas2018neural,lim2019pmnet,aberman2020skeleton,villegas2021contact}, which has advantages in intelligent perception and stable inference, becomes a new research trend. The previous learning-based methods utilize a {\it full-motion mapping} structure, which decodes joint rotations of the target skeleton as outputs, with joint positions~\cite{villegas2018neural,lim2019pmnet,villegas2021contact} or joint rotations~\cite{aberman2020skeleton} as inputs. However, due to the gap between the Cartesian coordinate space and the rotation space, the full joint position encoding unavoidably introduces motion distortion. Meanwhile, the full joint rotation encoding always leads to discontinuity in the rotation space~\cite{zhou2019continuity,aberman2020skeleton}.

In animation, we observe that artists usually copy the motion of the source character, and then manually modify it to preserve motion semantics and avoid translation artifacts, e.g., interpenetration, during motion reuse in new characters. Inspired by this observation, we design a \textbf{R}esidual \textbf{RET}argeting network (R$^2$ET) with a {\it residual} structure for motion retargeting. This structure takes the source motion as initialization and involves neural networks to imitate the modifications from animators, as illustrated in Figure~\ref{fig:1}. With this design, the coherence of the source motion is well maintained, and the search space for retargeting solutions during training is effectively reduced.

The key to achieving physically plausible single-body motion retargeting is to understand two main differences between the source and target characters, 1) the differences in bone length ratio; 2) the differences in body shape geometry. To reach this goal, we explore two modification modules, i.e., the skeleton-aware module and the shape-aware module, to perceive the two differences.

On the skeleton level, the {\it skeleton-aware} module takes the skeleton configurations as input to assist the transfer of the source motion semantics, such as arm folding and hand clapping, to the target character. To overcome the lack of paired and semantics-correct ground truth, we directly take the supervision signal from the input source motion. The motion semantics is explicitly modeled as a normalized Distance Matrix (DM) of skeleton joints. Accordingly, the semantics preservation is achieved by aligning the DM between the source and target motions (Figure~\ref{fig:1}-{\it Semantics}).

On the shape geometry level, the {\it shape-aware} module senses the compatibility between the target character mesh and the skeleton adjusted after motion semantics preservation to avoid interpenetration and contact-missing. To train the module end-to-end, we introduce two voxelized Distance Fields, i.e., the Repulsive Distance Field (RDF) and the Attractive Distance Field (ADF) (Figure~\ref{fig:1}-{\it Geometry}), as the measurement tools for interpenetration and contact. We sample the distance of the query vertices on the target character mesh to the body surface in these two fields to estimate the degree of interpenetration and contact. With this manner, the whole process is differentiable during training.

In practice, we find there always exists a contradiction between the preservation of motion semantics and the avoidance of interpenetration. We, therefore, propose a {\it balancing gate} to make a trade-off between the skeleton-level and geometry-level modifications by learning an adjusting weight. By leaving the weight to the user, our R$^2$ET also accepts interactive fine control from users.

With the above main designs, our R$^2$ET preserves the motion semantics of the source character and avoids interpenetration and contact-missing issues in a single-pass without post-processing. We evaluate our method on various complex motion sequences and a wide range of character geometries from skinny to bulky. The qualitative and quantitative results show that our R$^2$ET outperforms the existing learning-based methods by a large margin. The contributions of this work are summarized in three-fold:
\begin{compactitem}
\item A novel residual network structure is proposed for neural motion retargeting, which involves a skeleton-aware modification module, a shape-aware modification module, and a balancing gate.
\item A normalized joint Distance Matrix is presented to guide the training of the skeleton-aware module for explicit motion semantics modeling, and two Distance Fields are introduced to achieve differentiable pose adjustment learning.
\item Extensive experiments on the Mixamo \cite{mixamo} dataset demonstrate that our R$^2$ET achieves the state-of-the-art performance qualitatively and quantitatively. 
\end{compactitem}

\vspace{-.5mm}
\section{Related Work}
\vspace{-1.5mm}

\begin{figure*}[t]
\vspace{-.5cm}
\setlength{\abovecaptionskip}{-.2cm}
\setlength{\belowcaptionskip}{-.5cm}
\begin{center}
   \includegraphics[width=0.95\linewidth]{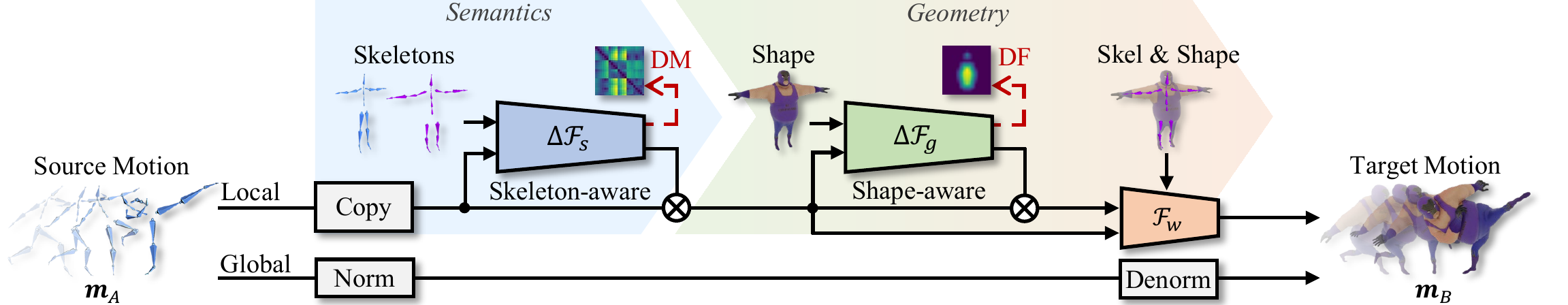}
\end{center}
   \caption{Overview of the proposed network R$^2$ET, which has three decoupled modules, i.e., the skeleton-aware module $\Delta \mathcal{F}_{s}$, the shape-aware module $\Delta \mathcal{F}_{g}$, and the balancing gate $\mathcal{F}_{w}$. The Distance Matrix (DM) and the Distance Field (DF) are two types of distance measurements that guide the network to learn the information of semantics and geometry.}
\label{fig:2}
\end{figure*}

\vspace{1mm}
\noindent\textbf{Motion Retargeting.} Motion retargeting is pioneered by \cite{gleicher1998retargetting}, which identifies features of the source motions as kinematic constraints and solves the space-time optimization problem. Following it, many optimization-based motion retargeting methods were proposed successively by introducing specific constraints, e.g., dynamics constraints \cite{tak2005physically}, inverse kinematics \cite{lee1999hierarchical}, joint angle constraints \cite{choi2000online}, Euclidean distance \cite{bernardin2017normalized}, and trajectory constraints \cite{feng2012automating}. Recently, there has been a surge of interest in studying deep-learning-based motion retargeting. Jang \textit{et al}. \cite{jang2018variational} used a deep autoencoder combining the DC-IGN \cite{kulkarni2015deep} and the U-Net \cite{ronneberger2015u} to generate human motions. Villegas \textit{et al}. \cite{villegas2018neural} trained a Neural Kinematic Network for unsupervised motion retargeting. Aberman \textit{et al}. \cite{aberman2020skeleton} proposed a Skeleton-Aware Network for retargeting motion between skeletons with different topologies. Lim \textit{et al}. \cite{lim2019pmnet} developed a novel architecture which separately learns frame-by-frame poses and overall movement. Li \textit{et al}. \cite{li2022iterative} introduced an iterative method to yield retargeted motions based on a trained motion autoencoder. All these methods were performed on articulated skeletons while ignoring the shape geometry of the characters. Thus, they are prone to produce unrealistic results on skinned motions. Villegas \textit{et al}. \cite{villegas2021contact} presented a latent-space optimization method for skinned motions to preserve self-contacts and prevent interpenetration, but this post-processing method is cumbersome and is unsuitable for a stable real-time system. In contrast to the above methods, we propose a new network R$^2$ET for skeletal and skinned motion retargeting with respect to motion semantics and character shapes, which can be trained in an end-to-end fashion.

\vspace{1mm}
\noindent\textbf{Geometry-aware Motion Modeling.} There is tremendous work on learning geometry-aware deep motion representations \cite{rempe2020contact}. Gomes \textit{et al}. \cite{gomes2021shape} leveraged the human body shape in the retargeting process while considering the physical constraints of the motion in the 2D image domain. Peng \textit{et al}. \cite{peng2021animatable} introduced the neural blend weight fields to reconstruct an animatable human model from a multi-view video. These works extract human motion from RGB-based videos or images, while our work looks at the motion retargeting of humanoid characters in 3D space. Jin \textit{et al}. \cite{jin2018aura} designed a volumetric mesh that surrounds a character’s skin to preserve the spatial relationships of humans. Basset \textit{et al}. \cite{basset2020contact} proposed an optimization-based method to deform the source shape in the desired pose using three energy functions. In this work, we designed two neural residual modules with distance-based losses to learn the motion semantics and the character geometrics for motion retargeting. Our method mainly copes with the animated characters with articulated skeletons, but can be also extended to retarget the motion from Skinned Multi-Person Linear Model (SMPL) \cite{loper2015smpl} estimated from RGB videos.

\vspace{-1mm}
\section{Method}
\vspace{-1.5mm}

Given the motion sequence $\bm{m}$ of the input character $A$ and the skeleton and mesh of a target character $B$ under rest-pose, two branches are involved to retarget the global root motion, i.e., velocities and rotations $\{\bm{v}^t\}^T_{t=1}$, $\bm{v} \in \mathbb{R}^4$, and the local joint rotation quaternions $\{\bm{q}^t\}^T_{t=1}$, $\bm{q}\in \mathbb{R}^{N \times 4}$, respectively, as Figure \ref{fig:2} shows. $N$ and $T$ indicate the number of joints and the sequence length. The time-index $t$ is ignored in the following for simplification. The global root movement is simply processed by normalizing and denormalizing with respect to the heights of the source and target character. The motion is translated framewisely. Inspired by the creation process in animation, we design a residual learning structure model R$^2$ET to achieve skeleton-aware and shape-aware local motion retargeting automatically. In particular, our R$^2$ET takes the source motion $\bm{q}_A$ as an initialization. A skeleton-aware modification module $\Delta \bm{q}_{s} = \Delta \mathcal{F}_{s}(\cdot)$ is introduced to maintain motion semantics, and a shape-aware modification module $\Delta \bm{q}_{g} = \Delta \mathcal{F}_{g}(\cdot)$ is involved to tackle the interpenetration and contact-missing issues. A balancing gate $\mathcal{F}_{w}$ is located at the end of this pipeline to balance the two motion modifications. The whole framework is then formulated as:
\begin{equation} \label{Eq.1}
\setlength{\abovedisplayskip}{4pt}
\setlength{\belowdisplayskip}{4pt}
\bm{q}_{B} = \mathcal{F}_{w}(~\bm{q}_{A}, ~\Delta\bm{q}_{s},~\Delta\bm{q}_{g}~),
\end{equation}
In contrast to \cite{villegas2018neural, lim2019pmnet} which take joint positions as input, we focus on retargeting the motions in the rotation space as~\cite{aberman2020skeleton}. The details of $\Delta \mathcal{F}_{s}$, $\Delta \mathcal{F}_{g}$ and $\mathcal{F}_{w}$ are introduced in Section \ref{sec:sem}, Section \ref{sec:geo} and Section \ref{sec:att}, respectively.

% \jw{Given the motion sequence $\bm{m}$ of the input character $A$, the goal of motion retargeting is to transfer this motion of $A$ to a character $B$ by considering the difference between these two characters.}
% \subsection{Overview}

\begin{figure*}[thb]
\vspace{-.5cm}
\setlength{\abovecaptionskip}{-.3cm}
\setlength{\belowcaptionskip}{-.4cm}
\begin{center}
   \includegraphics[width=0.95\linewidth]{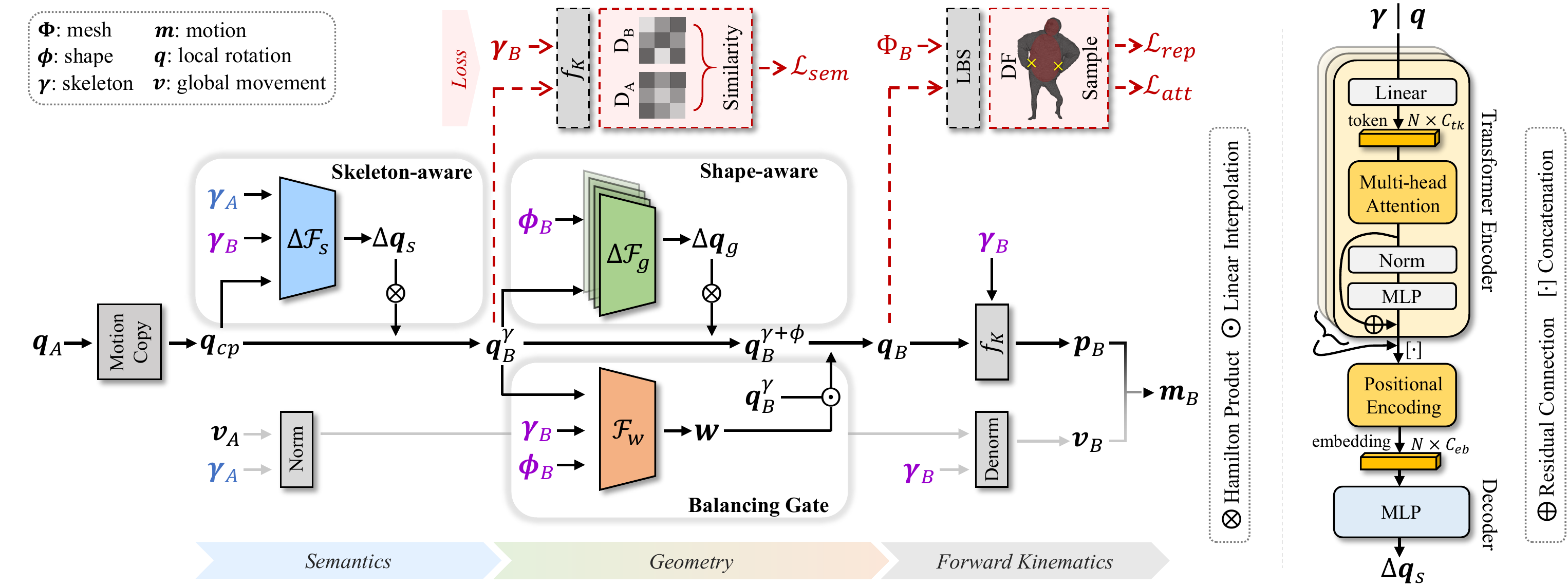}
\end{center}
  \caption{The detailed architectures of the proposed network R$^2$ET and the transformer-based skeleton-aware module.}
\label{fig:3}
\end{figure*}

\subsection{Base Losses}
\vspace{-1.5mm}
One of the challenges in neural motion retargeting is that there is always no paired ground truth as target motion supervision. Following \cite{lim2019pmnet}, we utilize the self-reconstruction principle and adversarial learning as basic training rules to train $ \Delta \mathcal{F}_{s}$, $\Delta \mathcal{F}_{g}$ and $\mathcal{F}_{w}$ in an unsupervised way.

During training, the self-reconstruction regularization is followed in two ways: 1) reconstructing the exact source motion in source character (see Section~\ref{sec:sem}), 2) minimizing the pose modifications in pose adjustment (see Section~\ref{sec:geo}). To avoid the quaternion ambiguity and reduce the position error accumulation along the kinematic chain, the joint rotations and positions are reconstructed simultaneously. Accordingly, the reconstruction loss is defined as:
\begin{equation} \label{Eq.2}
\setlength{\abovedisplayskip}{6pt}
\setlength{\belowdisplayskip}{6pt}
\begin{aligned}
\mathcal{L}_{rec}(\bm{q}, \hat{\bm{q}}) = \big\lVert\bm{q} - \hat{\bm{q}}\big\rVert_{2}^{2}+\big\lVert f_{K}(\bm{q}, \bm{\gamma}) - f_{K}(\hat{\bm{q}}, \bm{\gamma}) \big\rVert_{2}^{2},
\end{aligned}
\end{equation}
where $\bm{q}$ is the input rotation and $\hat{\bm{q}}$ is the estimated one. $f_{K}$ is a Forward Kinematics (FK) layer \cite{villegas2018neural} that maps the local joint rotations to the global joint positions by referring the rest-pose configuration $\bm{\gamma} \in \mathbb{R}^{N \times 3}$.

To achieve realistic motion retargeting, a discriminator $\delta (\cdot)$ is introduced to differentiate the translated motion sequences from the genuine ones. A motion discrimination loss is designed based on the adversarial training~\cite{goodfellow2020generative}:

\vspace{-2mm}
\begin{equation} \label{Eq.3}
\setlength{\abovedisplayskip}{0pt}
\setlength{\belowdisplayskip}{0pt}
\begin{aligned}
\mathcal{L}_{adv}(\hat{\bm{q}}) = &~\mathbb{E}_{\bm{m}\sim p_{real}}\big[\log \delta(\bm{m})\big]~+ \\
&~\mathbb{E}_{\bm{m}\sim p(\hat{\bm{q}})}\big[\log\big(1-\delta(\bm{m})\big)\big],
\end{aligned}
\end{equation}
in which $p(\cdot)$ represents the distribution of the {\it real} motions or {\it fake} retargeted motions controlled by $\hat{\bm{q}}$.

Besides, a rotation constraint loss \cite{villegas2018neural} is introduced to limit the $y$-axis Euler angles within a range, which avoids excessive joint twisting:

\vspace{-3mm}
\begin{equation} \label{Eq.4}
\setlength{\abovedisplayskip}{-10pt}
\setlength{\belowdisplayskip}{-5pt}
\begin{aligned}
\mathcal{L}_{rot}(\hat{\bm{q}}) = \big\lVert\max\big( \bm{0},\,\left|\epsilon_{y}(\hat{\bm{q}})\right|-\alpha\big)\big\lVert_{2}^{2}.
\end{aligned}
\end{equation}
The function $\epsilon_y(\cdot)$ converts the input quaternion to the Euler angle of $y$-axis, and $\alpha$ is the angle limitation bound. $\max(\cdot)$ is an element-wise function that returns the maximum number between the two inputs.

With these definitions, the base loss is thereafter defined by the weighted summation of the above losses as follows:
\begin{equation} \label{Eq.5}
\setlength{\abovedisplayskip}{5pt}
\setlength{\belowdisplayskip}{5pt}
\begin{aligned}
\mathcal{L}_{base}(\bm{q}, \hat{\bm{q}}) = \mathcal{L}_{rec} + \lambda \mathcal{L}_{adv} + \mu \mathcal{L}_{rot},
\end{aligned}
\end{equation}
where $\lambda$ and $\mu$ are the loss balancing factors.

\subsection{Skeleton-aware Module}
\label{sec:sem}
\vspace{-1.5mm}
The motion residual design can help maintain the motion coherence from the source character and meanwhile provides a good initialization for motion translation. However, owing to the differences of bone length and skeleton proportion between source and target characters, motion copy $\bm{q}_{cp}=\bm{q}_A$ may ignore the source motion semantics. In this section, we will introduce the skeleton-aware module $ \Delta \mathcal{F}_{s}$ for motion semantics preservation.

The skeleton-aware module takes the rest-pose skeletons $\bm{\gamma}_A$ and $\bm{\gamma}_B$ as well as the copied motion $\bm{q}_{cp}$ as input, and outputs the semantics-oriented quaternion modification $ \Delta \bm{q}_{s} \in \mathbb{R} ^{N \times 4}$. With this estimated modification, a Hamilton Product is applied to modify the motion copy $\bm{q}_{cp}$ and thereafter to preserve the motion semantics in the output $\bm{q}^{\gamma}_B$, namely:
% \vspace{-mm}
\begin{equation} \label{Eq.6}
\setlength{\abovedisplayskip}{3pt}
\setlength{\belowdisplayskip}{4pt}
\begin{aligned}
\bm{q}^{\gamma}_B & = \Delta \bm{q}_{s} \otimes \bm{q}_{cp}\\ & = \Delta \mathcal{F}_{s}(\bm{\gamma}_A, \bm{\gamma}_B, \bm{q}_{cp};~\bm{\theta}_s) \otimes \bm{q}_{cp},
\end{aligned}
\end{equation}
where $\theta_s$ indicates the parameter of module $\Delta \mathcal{F}_{s}$. 

\vspace{.5mm}
\noindent\textbf{Semantics Preservation.} There is no paired ground truth as strong semantics supervision. By utilizing the property of motion retargeting, we here take the supervision signals from the source motion. We model the motion semantics preservation as the maintaining of the source normalized pair-wise joint distances in the target character pose, frame by frame. We introduce a normalized Distance Matrix (DM) $\bm{D}\in \mathbb{R} ^{N \times N}$ to represent the motion semantics, i.e., pair-wise joint distances as shown in Figure~\ref{fig:4}. The columns of the matrix indicate the query joints, and the rows represent the reference joints. The element $d_{i,j}$ of $\bm{D}$ represents the Euclidean distance from query joint $i$ to reference joint $j$. We extract the pose DM from the source character and regard it as a supervision signal to guide the learning of target pose DM. With this design, a Semantics Similarity loss is then defined as:
\begin{equation} \label{Eq.7}
\setlength{\abovedisplayskip}{3pt}
\setlength{\belowdisplayskip}{3pt}
\begin{aligned}
\mathcal{L}_{sem} = \Bigg\lVert \eta\bigg(\frac{\bm{{D}}_{A}}{h_{A}}\bigg) - \eta\bigg(\frac{\bm{{D}}_{B}}{h_{B}}\bigg) \Bigg\rVert_{2}^{2},
\end{aligned}
\end{equation}
where $h$ is the height of the skeleton. $\eta(\cdot)$ is an $L1$ normalization performed on each row of the distance matrix. This normalization operation eliminates the difference of bone lengths and heights between the source and target skeletons.

To support the pair-wise joint relationship learning, we introduce a Transformer structure~\cite{vaswani2017attention}, whose attention mechanism is suitable for pair-wise learning, to build the skeleton-aware module. As the right of Figure~\ref{fig:3} shows, our Transformer-based structure consists of two Transformer encoders and one MLP decoder. The Transformer encoders process $\bm{\gamma}$ and $\bm{q}$ independently. In this process, $N$ joint features are treated as $N$ tokens with $C_{tk}$ channels, and they are encoded by an Multi-head Attention and a Layer Normalization operations. Then, the feature of $\bm{\gamma}$ and $\bm{q}$ are concatenated and position-encoded to obtain an embedding with $C_{eb}$ channels. In the end, an MLP is shared within $N$ joints to decode the rotation modifications $\Delta \bm{q}_{s}$ for these joints. With the loss and model structure introduced above, the skeleton-aware module can be trained by:

\vspace{-4mm}
\begin{equation} \label{Eq.8}
\vspace{-2mm}
\setlength{\abovedisplayskip}{-10pt}
\setlength{\belowdisplayskip}{-10pt}
\begin{aligned}
\min_{\bm{\theta}_s}~\mathcal{L}_{base}(\bm{q}_{cp}, \bm{q}^{\gamma}_B) + \nu \mathcal{L}_{sem},
\end{aligned}
\end{equation}
where $\nu$ is the loss balancing factor. The reconstruction loss is applied when the source and the target characters are the same. We sample the target character as the source one with a probability of 0.5.

\subsection{Shape-aware Module}
\label{sec:geo}
\vspace{-1.5mm}
We introduce a shape-aware module $\Delta \mathcal{F}_{g}$ in this section to ensure the retargeted skinned motion is interpenetration-free and contact-preserved, as illustrated in the middle part of Figure~\ref{fig:3}. The shape-aware module takes the shape information $\bm{\phi}$ of each body part in the target character as well as the $\Delta \mathcal{F}_{s}$-modified local joint rotation $\bm{q}^{\gamma}_{B}$ as input, and outputs the geometry-oriented quaternion modification $\Delta \bm{q}_{g} \in \mathbb{R} ^{N \times 4}$. $\bm{\phi} \in \mathbb{R}^{N \times 3}$ is represented by the edge-lengths of the body part bounding box corresponding to each joint in the rest-pose. As we observed, most of the interpenetration and contact-missing issues occur between the limbs and the main body. We, therefore, choose to only adjust the rotations 
 of four target limbs and introduce four MLPs to estimate rotation modifications for them independently. With the estimated rotation modifications, the adjusted joint rotation $\bm{q}^{\gamma+\phi}_{B}$ is then defined as:
\begin{equation} \label{Eq.9}
\setlength{\abovedisplayskip}{3pt}
\setlength{\belowdisplayskip}{3pt}
\begin{aligned}
\bm{q}^{\gamma+\phi}_B & = \Delta \bm{q}_{g} \otimes \bm{q}^{\gamma}_B \\
& = \Delta \mathcal{F}_{g}(\bm{\phi}_B, \bm{q}^{\gamma}_B;~\bm{\theta}_g) \otimes \bm{q}^{\gamma}_B,
\end{aligned} 
\end{equation}
where $\theta_g$ represents the parameter of module $\Delta \mathcal{F}_{g}$.

\begin{figure}[t]
\vspace{-.4cm}
\setlength{\abovecaptionskip}{-.3cm}
\setlength{\belowcaptionskip}{-.5cm}
\begin{center}
   \includegraphics[width=0.95\linewidth]{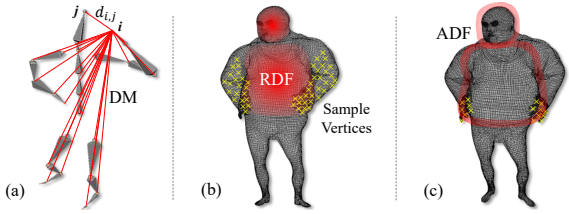}
\end{center}
   \caption{Illustration of the distance measurements. DM is the normalized Distance Matrix of the skeleton joints. RDF and ADF are the Distance Fields inside and outside the main body.}
\label{fig:4}
\end{figure}

\vspace{1mm}
\noindent\textbf{Penetration-free \& Contact-preserving.} To achieve differentiable pose adjustment learning in respect of mesh geometry, we introduce two truncated distance fields, the Repulsive Distance Field (RDF) $\psi_R$ inside the main body and the Attractive Distance Field (ADF) $\psi_A$ around the body surface. These two fields are illustrated in Figure~\ref{fig:4} (b,c). RDF assists the network to force the penetrating vertices to be apart from the interpenetration area, and the ADF attracts the near-contact vertices adjusted after motion semantics preservation to be close to the body surface. With the adjusted rotation $\bm{q}^{\gamma+\phi}_{B}$, we first deform the mesh vertice set of the target character $\Phi_B$ by applying Linear Blend Skinning (LBS). Hereafter, $\psi_R$ and $\psi_A$ are estimated by voxelizing the deformed target mesh. Each node on the voxel grid records its distance to the body surface from inside or outside, and we can thus measure the body-surface deviation of each query vertice $e$ on the deformed target mesh by interpolating four node distances on its surrounding tight voxel grid. With this mechanism, our model can be trained to handle the interpenetration and contact-missing problem in an end-to-end manner, and translate a source motion to a plausible target motion during inference without post-processing. These two fields are embedded into two losses, the Repulsive and Attractive loss, respectively to achieve end-to-end training, and they are defined as:
\begin{equation} \label{Eq.10}
\setlength{\abovedisplayskip}{3pt}
\setlength{\belowdisplayskip}{3pt}
\begin{aligned}
\mathcal{L}_{rep} = \frac{1}{N_{l}}\sum\limits_{e \in E_{l}} \psi_{R}(e),~~~~\mathcal{L}_{att} = \frac{1}{N_{h}}{\sum\limits_{e \in E_{h}} \psi_{A}(e)},
\end{aligned}
\end{equation}
where $E_l=\{e_i\}^{N_l}_{i=1}$ and $E_h=\{e_i\}^{N_h}_{i=1}$ are the vertices set of the deformed target mesh's limbs and hands, respectively. $N_l$ and $N_h$ are the corresponding numbers of vertices. $\psi(e)$ samples the $\psi$ value for each vertex $e$ in a differentiable way. $ \Delta \mathcal{F}_{g}$ consists of four independent networks corresponding to the four limbs of the character. Accordingly, these four networks are optimized by four related $\mathcal{L}^{E_l}_{rep}$ as:
% \vspace{-.2cm}
\begin{equation} \label{Eq.11}
\setlength{\abovedisplayskip}{1pt}
\setlength{\belowdisplayskip}{3pt}
\begin{aligned}
% \min_{\{\theta_g\}}~\mathcal{L}_{base}(\bm{q}^{\gamma}_B, \bm{q}^{\gamma+\phi}_B) + \kappa \sum\limits_{i=1}^{4} \mathcal{L}^{E^i_l}_{rep},
\min_{\bm{\theta}_g}~\mathcal{L}_{base}(\bm{q}^{\gamma}_B, \bm{q}^{\gamma+\phi}_B;~\bm{\theta}_g) + \kappa \sum\limits_{i=1}^{4} \mathcal{L}^{E^i_l}_{rep}(\ \cdotp\ ;~\bm{\theta}^i_g),
\end{aligned}
\vspace{-.2cm}
\end{equation}
% \vspace{-.2cm}
where $\kappa$ is the balancing hyper-parameter, $\bm{\theta}_g=[\bm{\theta}^i_g]_{i=1}^4$. As repulsing and attracting the mesh vertices simultaneously would cause unstable training convergence, we do not involve $\mathcal{L}_{att}$ here but leave it in the next Balancing module.

\begin{figure*}[t]
\vspace{-.5cm}
\setlength{\abovecaptionskip}{-.2cm}
\setlength{\belowcaptionskip}{-.4cm}
\begin{center}
   \includegraphics[width=1.0\linewidth]{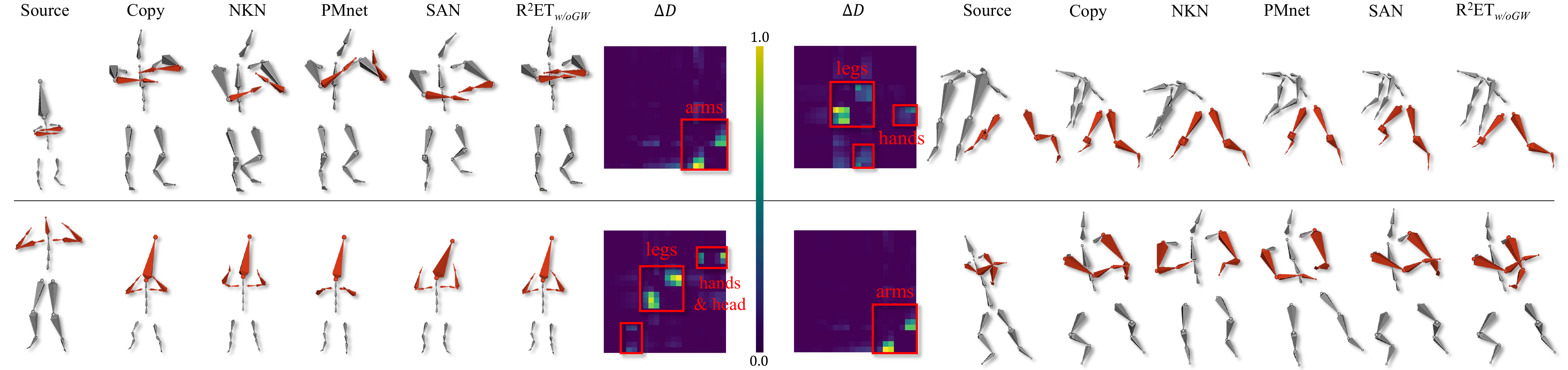}
\end{center}
   \caption{Qualitative results of skeletal motion retargeting. $\Delta D$ indicates the DM difference between the motion copy and our result.}
\label{fig:5}
\end{figure*}

\subsection{Balancing Gate}
\label{sec:att}
\vspace{-1.5mm}
In practice, it is challenging to learn the motion semantics preservation at the skeleton level and meanwhile train the network to tackle the interpenetration as well as contact-missing issues at the shape-geometry level. Let's take the bottom of Figure~\ref{fig:1} as an instance. When retargeting motion from a thin character to an obese character, if only the relative positions of the joints are maintained, it will inevitably lead to interpenetration. On the other side, if only the target shape is considered, the retargeted motion may lose motion semantics. To overcome this obstacle, we introduce an additional MLP module $\mathcal{F}_{w}$ to balance the influence between the two modifications $\Delta\bm{q}_s$ and $\Delta\bm{q}_g$ by a learned balancing factor $\bm{w} \in \mathbb{R}^{N}$. This balancing process is achieved by a linear interpolation between $q^{\gamma}_B$ and $q^{\gamma+\phi}_B$:
\begin{equation} \label{Eq.12}
\setlength{\abovedisplayskip}{3pt}
\setlength{\belowdisplayskip}{5pt}
\begin{aligned}
\bm{q}_{B} = (\bm{1}-\bm{w})\cdot\bm{q}^{\gamma}_B + \bm{w}\cdot\bm{q}^{\gamma+\phi}_B.
\end{aligned}
\end{equation}
in which $\bm{w} = \mathcal{F}_{w}(\bm{\gamma}_B, \bm{\phi}_B, \bm{q}^{\gamma}_B; \theta_w)$. $\theta_w$ indicates the parameters of $\mathcal{F}_{w}$. Each element of $w$ is ranged from 0 to 1. The symbol ``$\cdot$" indicates scaling each row of $\bm{q}$ by each element of $\bm{w}$. By leaving the vector $\bm{w}$ to the user, we can also manually adjust its value at each joint to finely control the retargeted results. To reach an optimized balancing, $\mathcal{F}_{w}$ can be trained by:
\begin{equation} \label{Eq.13}
\setlength{\abovedisplayskip}{4pt}
\setlength{\belowdisplayskip}{2pt}
\begin{aligned}
\min_{\bm{\theta}_w}~\mathcal{L}_{base}(\bm{q}^{\gamma+\phi}_B, \bm{q}_B) + \kappa\mathcal{L}_{rep} + \iota \mathcal{L}_{att} + \tau \mathcal{L}_{reg},
\end{aligned}
\end{equation}
where $\kappa$, $\iota$, and $\tau$ are hyper-parameters. $\mathcal{L}_{reg}$ is a $L2$ regularization loss for $w$.

\vspace{-1mm}
\section{Experiments}
\vspace{-1mm}
\noindent\textbf{Datasets.} We evaluate our R$^2$ET on the Mixamo dataset \cite{mixamo}, which is an animation repository performed by multiple 3D virtual characters with different skeletons and shapes. For training, we collect 1952 non-overlapping motion sequences of seven characters and randomly sample 60 frames from each sequence. For testing, we collect 800 motion sequences of 11 characters and each sequence has 120 frames. We have unseen character (UC), unseen motion (UM), seen character (SC), and seen motion (SM) so that four splits UC+UM, UC+SM, SC+UM, SC+SM are considered in the experiment. Around 3/4 of the test samples are unseen. The Mixamo dataset does not provide clean Ground Truth (GT): many of the motions may have interpenetration or contact-missing issues which makes geometry learning challenging. For a fair comparison, we follow the spirit of \cite{villegas2018neural} to implement experiments.

% The token channels $C_{tk}$ of the $\gamma$ and $q_{cp}$ are 64. The embedding channels $C_{eb}$ of them are 128 and 256, respectively. The number of attention heads for encoding $\gamma$ is 2, and that for encoding $q_{cp}$ is 4. To train the skeleton-aware module, the learning rate is set as 0.001, the number of training epochs is set as 30 and the batch size is 32. To train the shape-aware module and the balancing gate, the learning rate is set as 0.0001, the number of training epochs is set as 50 and the batch size is 16.

\vspace{1mm}
\noindent\textbf{Implementation details.} We select $N=22$ joints for each character in our experiments. The hyper-parameters $\lambda$, $\mu$, $\nu$, $\kappa$, $\iota$, and $\tau$ in loss functions are set as 2.0, 10.0, 100.0, 0.5, 0.5, and 0.005, respectively. The margin factor $\alpha$ in the Rotation Constraint loss is defined as 100. We implement our model based on the PyTorch framework~\cite{paszke2019pytorch} and apply the Adam optimizer \cite{kingma2015adam} to train the network. The training process is divided into two stages: firstly train the skeleton-aware module independently, then freeze its parameters and train the shape-aware module and the balancing gate. Please see Sup. Mat. for more details.

\vspace{1mm}
\noindent\textbf{Evaluation metrics.} We evaluate the results from two aspects: skeleton and geometry levels. For skeletal motions, the metric Mean Square Error (MSE) and local MSE are to measure how close the retargeted joint positions are to the GT provided in the Mixamo. The local MSE is calculated when the root of the retargeted motion is aligned with that of the GT. The joint position error is normalized by the character height~\cite{villegas2018neural}. The geometric evaluation includes the measurement of interpenetration and self-contact. Interpenetration is scored by the ratio of the number of penetrated limb vertices to the total number of limb vertices in each frame. Self-contact is measured by the average distance from the vertex of the hands to the surface of the body. Additionally, we also implement qualitative analysis and user study to verify the advance of our method. The methods included for comparison are recent deep-learning-based methods NKN \cite{villegas2018neural}, PMnet \cite{lim2019pmnet}, and SAN \cite{aberman2020skeleton}.

\begin{figure*}[t]
\vspace{-.5cm}
\setlength{\abovecaptionskip}{-.2cm}
\setlength{\belowcaptionskip}{-.6cm}
\begin{center}
   \includegraphics[width=0.95\linewidth]{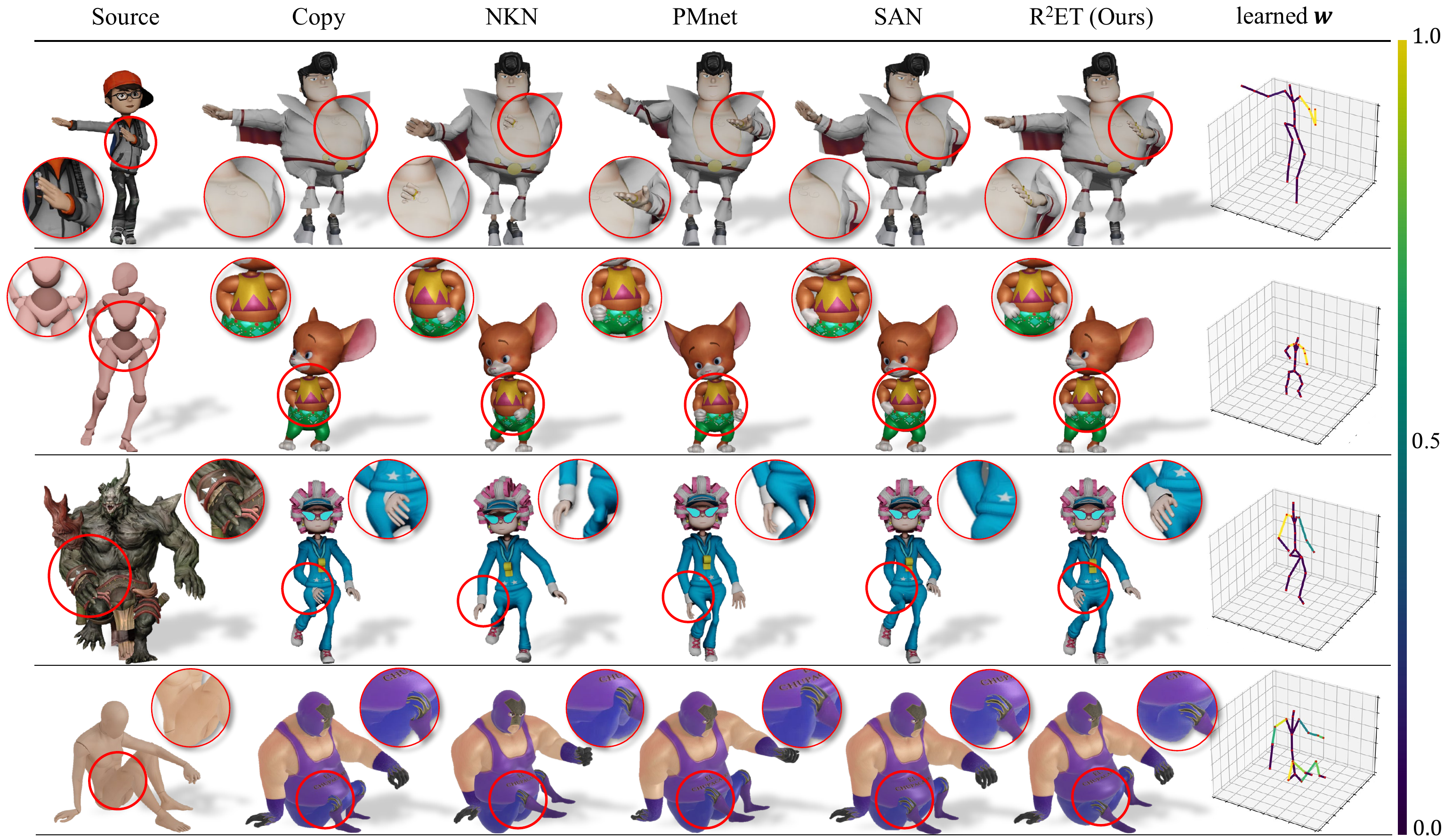}
\end{center}
   \caption{Qualitative results of skinned motion retargeting. $\bm{w}$ is the adjusting weight that learned by the balancing gate.}
\label{fig:6}
\end{figure*}

\vspace{-1mm}
\subsection{Qualitative Results}
\vspace{-1mm}
\noindent\textbf{Semantics.} Figure \ref{fig:5} visualizes the effect of the skeleton-aware module of R$^2$ET on motion semantics preservation in skeletal motion retargeting. R$^2$ET$_{w/oGW}$ means our method with only the Skeleton-aware Module equipped. We retarget the motion of the bones among small, medium, and large skeletons with different bone length ratios. The skeleton-aware module can well preserve the motion semantics according to the characteristics of the target character's skeleton. For example, in the top-left row of Figure \ref{fig:5}, the "arm folding" pose is retargeted from a small character to a large one, but the existing methods cannot accurately preserve the semantics, and the result of motion copy is more like "hand clapping". Our R$^2$ET can perceive the key differences between these two skeletons, i.e., the bone length ratio of the arms to the forearms, and adaptively adjust the copied rotations to generate a reasonable retargeted motion. The following three cases further demonstrate that our method, driven by the Semantics Similarity Loss, can well perceive the skeleton motion semantics and translate it between skeletons with large differences.

\vspace{1mm}
\noindent\textbf{Geometry.} Figure \ref{fig:6} shows the results of skinned motion retargeting among characters with different shapes. The target characters of the last two rows are unseen characters. The existing methods barely consider the shape geometry of target characters and their results suffer from severe interpenetration and contact-missing issues. Our results, which are based on the semantics-preserved motion and adjusted by the shape-aware module as well as the balancing gate, can well reduce these implausible issues while maintaining motion semantics as much as possible without the post-processing. Figure \ref{fig:6} also visualizes the $\bm{w}$ of each joint predicted by the balancing gate. The predicted $\bm{w}$ have higher responses on the joints whose succeeding body parts are interpenetrated, and have lower responses when the motion semantics of the corresponding parts need to be preserved.

\begin{figure}[t]
% \vspace{-1cm}
\setlength{\abovecaptionskip}{-.3cm}
\setlength{\belowcaptionskip}{-.6cm}
\begin{center}
   \includegraphics[width=1.0\linewidth]{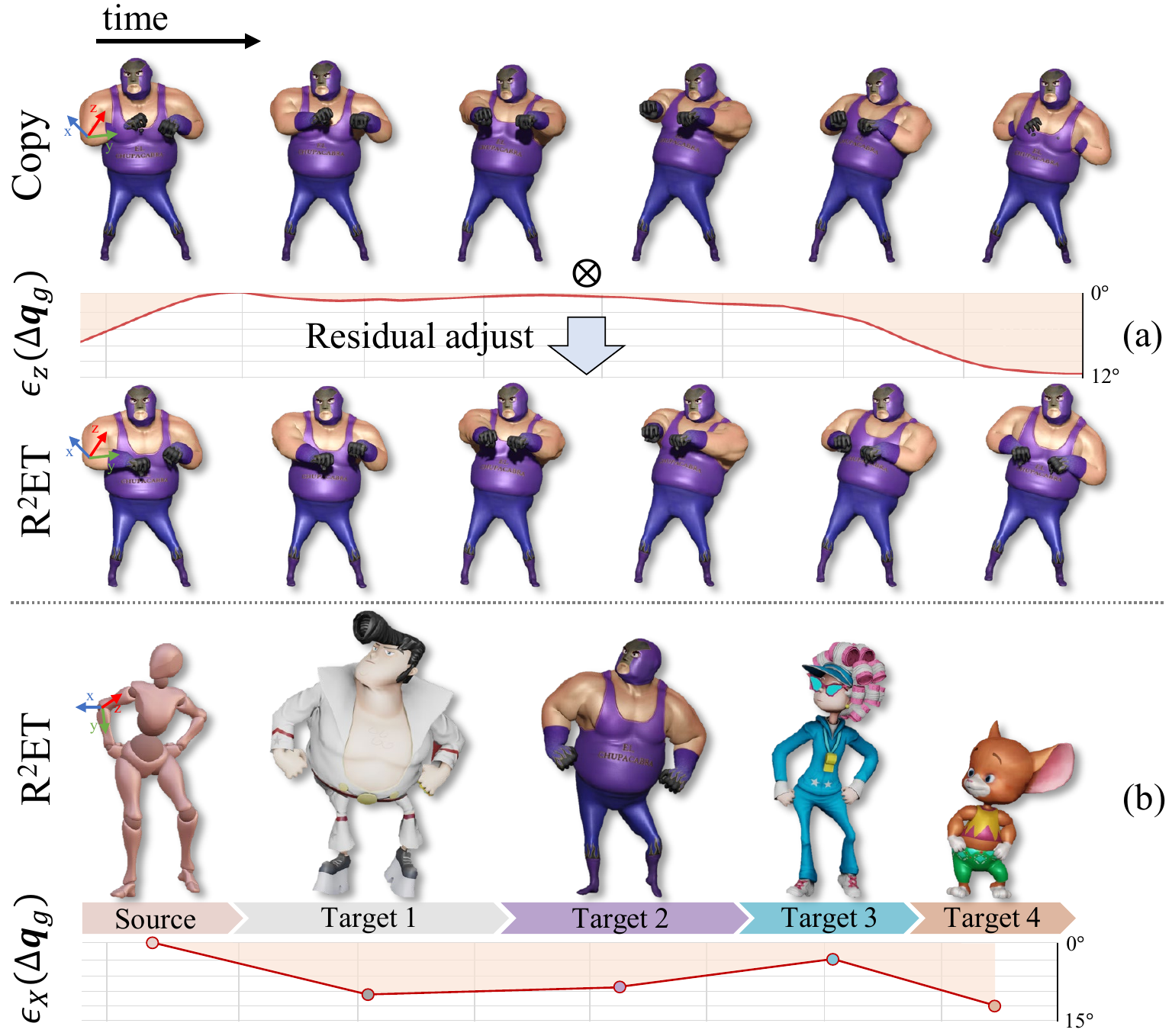}
\end{center}
   \caption{(a) The change of the geometry-oriented modification $\Delta \bm{q}_{g}$ of a motion sequence on time domain. (b) Our results of retargeting one source to multi-targets.}
\label{fig:7}
\end{figure}

Figure \ref{fig:7} (a) shows the change of the geometry-oriented modification $\Delta \bm{q}_{g}$ with time of a motion sequence in the Mixamo dataset. The target 1,2,3 are unseen characters. The vector $\Delta \bm{q}_{g}$ is converted to the average z-axis Euler angle value of two arms for simple illustration. Our shape-aware module can accurately perceive the poses with interpenetration problems as the change of time and apply reasonable adjustments to them. For the poses that are not suffering from interpenetration, our R$^2$ET hardly adjusts them, so as to keep the original motion semantics as much as possible. At the same time, the modification changes smoothly as time goes on, which ensures the coherence and naturalness of the retargeted motion. Figure \ref{fig:7} (b) shows our results of retargeting one source motion to multiple targets. For characters with different body shapes, our R$^2$ET can sensitively perceive their geometries and make precise adjustments to the source motion. Overall, Figure \ref{fig:7} demonstrates that our R$^2$ET is robust to a variety of poses and characters.

\begin{figure}[t]
% \vspace{-.4cm}
\setlength{\abovecaptionskip}{-.4cm}
\setlength{\belowcaptionskip}{-.6cm}
\begin{center}
   \includegraphics[width=0.95\linewidth]{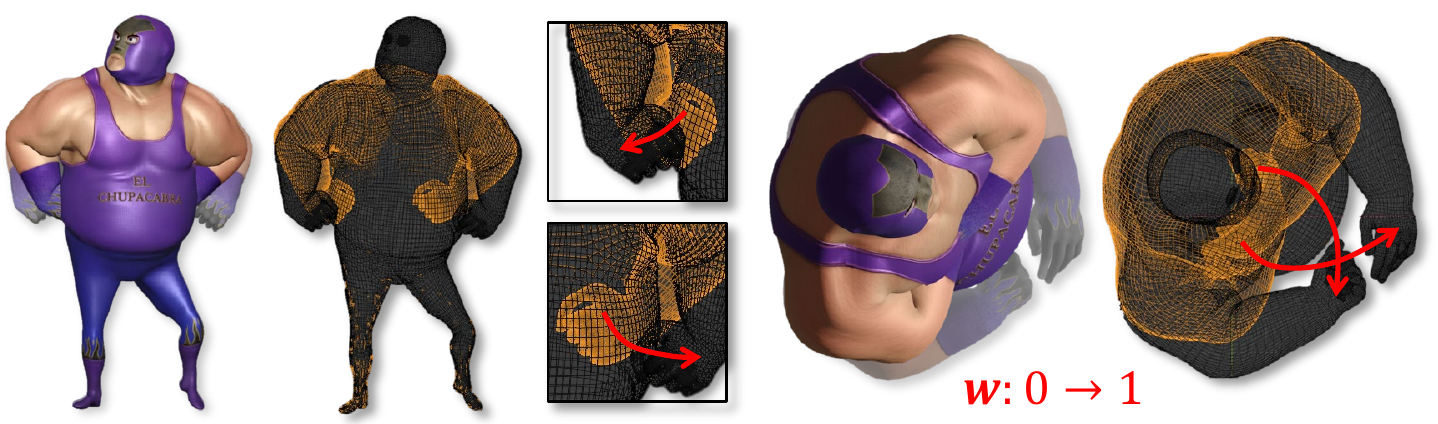}
\end{center}
   \caption{Manually adjusting the balancing gate $\bm{w}$ can obtain smooth motion adjustment.}
\label{fig:8}
\end{figure}

An automatic algorithm can provide visual results that follow the pre-defined learning constraints designed by engineers, such as the avoidance of interpenetration and contact-missing, but cannot always satisfy the aesthetic needs of the animators. Thanks to the flexible balancing gate, our R$^2$ET overcomes this drawback as Figure \ref{fig:8} shows. By gradually scaling $\bm{w}$, we can get results that vary smoothly from the arm pose that preserves the motion semantics to the one that avoids interpenetration, thereby interactively selecting the visually best results.

\vspace{-1mm}
\subsection{Quantitative Results}
\vspace{-1mm}
\noindent\textbf{Comparison with the state-of-the-arts.} Table~\ref{table:1} shows the comparison between our method and the state-of-the-arts. Considering that the Mixamo dataset may create a new character with an archived motion by using motion copy, the ``Copy'' has the lowest MSE and local MSE. However, this does not mean that the motion copy is the best choice (See Figure \ref{fig:5} and Figure \ref{fig:6}). We here just treat MSE as an auxiliary reference metric for comparison. Compared with NKN, PMnet, and SAN which focus on skeletal motion retargeting, our R$^2$ET$_{w/oGW}$ reduces the MSE by 87\% (0.297 vs 2.298), 63\% (0.297 vs 0.806), and 7\% (0.297 vs 0.321), respectively. The MSE of the PMnet with rotation-input (PMnet*) is lower than the PMnet with position-input but is also worse than ours. The above results show that our method can well reconstruct the source motion while adjusting the local motion according to the skeletal configurations to make it more in line with the motion semantics.

\begin{table}[t]
\vspace{-.4cm}
\setlength{\abovecaptionskip}{-.1mm}
\setlength{\belowcaptionskip}{-.15cm}
\centering
\caption{Comparison with the state-of-the-arts. MSE$^{lc}$ is the local MSE. R$^2$ET$_{w/oGW}$ is the model with the skeleton-aware module only. R$^2$ET$_{w/oW}$ is the model without the balancing gate. Copy\dag{} is the motion copy without the global motion normalization.}
\small{
\begin{tabular}{ l c | c c | c c} 
		\toprule[1.5pt]
		\textbf{Methods} & \textbf{Inp.} & \textbf{MSE}$_{\downarrow}$ & \textbf{MSE$^{lc}$}$_{\downarrow}$ & \textbf{Pen.}$^{\%}_{\downarrow}$ & \textbf{Con.}$^{cm}_{\downarrow}$\\
             \midrule[0.5pt]
            GT & - & - & - & 9.02 & 4.92  \\
		\midrule[0.5pt]
		NKN \cite{villegas2018neural} & \multirow{2}{*}{Pos.} & 2.298 & 0.575 & 8.96 & 4.42  \\
		PMnet \cite{lim2019pmnet} &  & 0.806 & 0.281 & 7.11 & 14.7  \\
		\midrule[0.5pt]
            Copy & \multirow{4}{*}{Rot.} & \underline{0.267} & \underline{0.060} & 9.23 & 4.95  \\
            Copy\dag & & 3.087 & 0.060 & 9.23 & 4.95  \\
		SAN \cite{aberman2020skeleton} &  & 0.321 & 0.118 & 8.91 & 4.86  \\
            PMnet* &  & 0.374 & 0.120 & 9.03 & 5.24  \\
		\midrule[0.5pt]
		R$^2$ET$_{w/oGW}$ & \multirow{3}{*}{Rot.} & \textbf{0.297} & \textbf{0.094} & 9.09 & 4.93  \\
            R$^2$ET$_{w/oW}$ & & 0.378 & 0.178 & \textbf{4.68} & 5.31   \\
            R$^2$ET (Ours) & & 0.318 & 0.116 & 5.94 & \textbf{3.57}   \\
		\bottomrule[1.5pt]
	\end{tabular}
 }
    \label{table:1}
% \end{table}
\vspace{2mm}
% \begin{table}[t]
% \setlength{\abovecaptionskip}{-.01cm}
% \setlength{\belowcaptionskip}{-.1cm}
\centering
\caption{Ranking results of the user study. We invite 100 users to compare our retargeting results to that of the recent methods from three aspects, i.e., overall quality (Q), semantics preservation (S), and motion details (D).}
\small{
\begin{tabular}{ l | c c c | c c c} 
\toprule[1.5pt]
   \textbf{\multirow{2}{*}{Methods}} & \multicolumn{3}{c}{\textbf{Skeletal Motion}} & \multicolumn{3}{c}{\textbf{Skinned Motion}} \\
    \cmidrule{2-7}
     % & Quality & Semantics & Detail & Quality & Semantics & Detail \\
     & Q$_{\downarrow}$ & S$_{\downarrow}$ & D$_{\downarrow}$ & Q$_{\downarrow}$ & S$_{\downarrow}$ & D$_{\downarrow}$ \\
    \midrule[0.5pt]
    Copy & 1.88 & 1.83 & 1.84 & 1.84 & 1.84 & 1.93 \\
    NKN \cite{villegas2018neural} & 3.37 & 3.45 & 3.40 & 3.44 & 3.44 & 3.42 \\
    PMnet \cite{lim2019pmnet} & 3.06 & 3.06 & 3.06 & 3.10 & 3.07 & 3.00 \\
    R$^2$ET (Ours) & \textbf{1.69} & \textbf{1.67} & \textbf{1.70} & \textbf{1.63} & \textbf{1.65} & \textbf{1.64}\\
            
\bottomrule[1.5pt]
\end{tabular}
}
\label{table:2}
\vspace{-.4cm}
\end{table}

As Table \ref{table:1} shows, the GT of the Mixamo dataset bears the interpenetration and contact-missing issues. Our R$^2$ET$_{w/oW}$, with the help of the shape-aware module, can perceive the geometry of characters and reduce the interpenetration effectively. Compared with the GT, our R$^2$ET$_{w/oW}$ reduces the penetration rate by more than 48\% (4.68 vs 9.02). Without the balancing gate equipped, the contact can not be well maintained while reducing interpenetration. Our full model, i.e., R$^2$ET, reaches a good balance among these three quantitative metrics and also obtains the best qualitative visualization results (see Figure \ref{fig:6}). More results about the seen/unseen splits can be seen in Sup. Mat.

\begin{figure}[t]
\vspace{-.6cm}
\setlength{\abovecaptionskip}{-.01mm}
\centering
\includegraphics[width=1.0\linewidth]{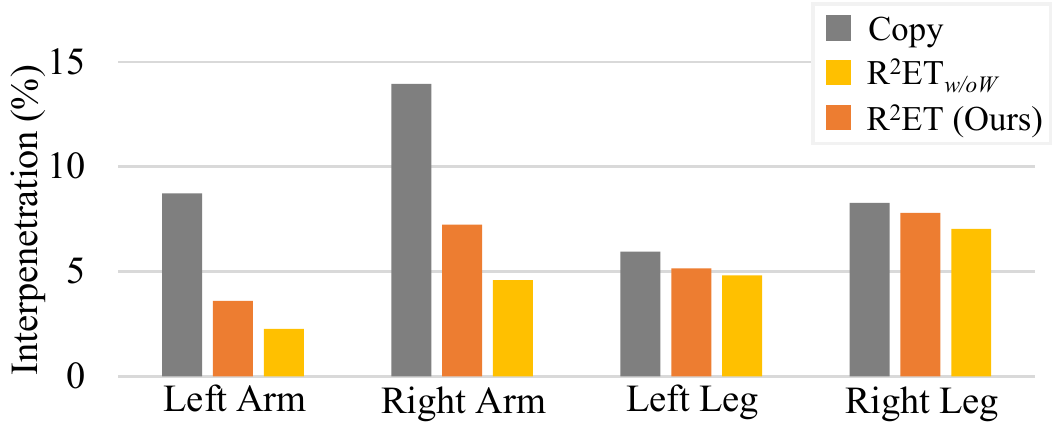}
\captionof{figure}{Comparison of the interp. rates of different limbs.}
\label{fig:9}
\vspace{2mm}
\centering
\includegraphics[width=1.0\linewidth]{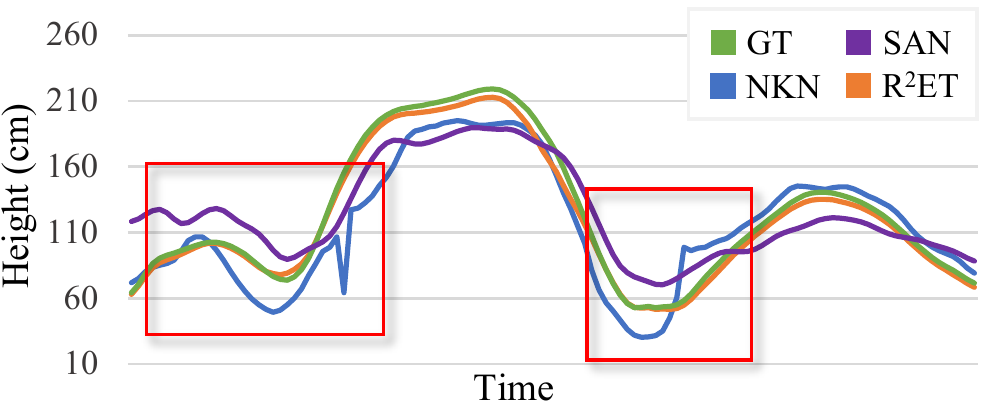}
\captionof{figure}{The change of the left-hand end-effector's height of a retargeted motion on time domain.}
\label{fig:10}
\vspace{-.4cm}
\end{figure}

Figure \ref{fig:9} shows the comparisons of the penetration rates of the character's four limbs between the motion copy and our models. The interpenetration issues mainly occur between the arms and the body of the character, and our models can significantly reduce their penetration rates.

The Contact-aware Model in \cite{villegas2021contact} focuses on skinned motion retargeting and can also effectively reduce interpenetration and preserve self-contacts. However, unlike our R$^2$ET, the Contact-aware Model adopts a post-processing method to optimize the latent space of motion feature, which may not generate plausible results in a single inference pass, and it may result in unstable real-time inference. Therefore, our method is more efficient and easier to use.

Figure \ref{fig:10} shows the change of the end-effector's height of a retargeted motion on time domain. Compared to the NKN and SAN that are based on the full-motion mapping structure, our R$^2$ET with the residual structure can obtain smooth and stable retargeted motion in time series.

\vspace{-1mm}
\subsection{User Study}
\vspace{-1mm}
% \noindent\textbf{User study.}
We conduct a user study to evaluate the performance of our R$^2$ET against the relevant methods NKN, PMnet, and motion copy. We invited 100 users and gave them six skeletal action videos and seven shape action videos in total to evaluate. Each video includes one source motion and four anonymous results. We ask users to rank the four results in three aspects: overall quality (Q), semantics preservation (S), and motion details (D). After excluding abnormal questionnaires, we collect 3120 ranking comparisons statistics in total and the average rank of the methods is summarized in Table \ref{table:2}. For skeletal motion, our method ranks 1.68 on average. For skinned motion, our method ranks 1.64 on average. In general, more than 71.2\% of users prefer the retargeting results of our method.
% \vspace{-2mm}

\vspace{-1mm}
\section{Conclusions}
\vspace{-1mm}
A novel network R$^2$ET with residual structure is proposed for neural motion retargeting. In R$^2$ET, two motion modification modules are explored to assist in generating plausible target motion. The skeleton-aware module adjusts the input motion to preserve source motion semantics in the target character. The shape-aware module senses the compatibility between the target shape and semantics-preserved pose to avoid interpenetration and contact-missing. Besides, a balancing gate is designed to make a trade-off between the skeleton-level and geometry-level modifications by learning an adjusting weight. With the help of two distance-based measurements, R$^2$ET is trained end-to-end. Extensive experiments on the Mixamo dataset demonstrate that our R$^2$ET achieves the state-of-the-art motion retargeting performance. It provides a good balance between the preservation of motion semantics and the reduction of interpenetration and contact-missing without post-processing.

\noindent\textbf{Limitations.} One potential drawback lies in the noisy motion data of the adopted Mixamo dataset. We endeavor to reduce noise interference for future work. Foot contact is not our focus but can be tackled by a method in \cite{aberman2020skeleton}.

\noindent\textbf{Acknowledgements.} This work was supported by the fund of Tencent AI Lab RBFR2022012, and the Natural Science Fund for Distinguished Young Scholars of Hubei Province under Grant 2022CFA075.

\clearpage
%%%%%%%%% REFERENCES
{\small
\bibliographystyle{ieee_fullname}
% \bibliography{egbib}

\begin{thebibliography}{10}\itemsep=-1pt

\bibitem{aberman2020skeleton}
Kfir Aberman, Peizhuo Li, Dani Lischinski, Olga Sorkine-Hornung, Daniel
  Cohen-Or, and Baoquan Chen.
\newblock Skeleton-aware networks for deep motion retargeting.
\newblock {\em ACM Transactions on Graphics}, 39(4):62--1, 2020.

\bibitem{mixamo}
Adobe.
\newblock Mixamo.
\newblock \url{https://www.mixamo.com/}.

\bibitem{basset2020contact}
Jean Basset, Stefanie Wuhrer, Edmond Boyer, and Franck Multon.
\newblock Contact preserving shape transfer: Retargeting motion from one shape
  to another.
\newblock {\em Computers \& Graphics}, 89:11--23, 2020.

\bibitem{bernardin2017normalized}
Antonin Bernardin, Ludovic Hoyet, Antonio Mucherino, Douglas Gon{\c{c}}alves,
  and Franck Multon.
\newblock Normalized euclidean distance matrices for human motion retargeting.
\newblock In {\em Proceedings of the Tenth International Conference on Motion
  in Games}, pages 1--6, 2017.

\bibitem{choi2000online}
Kwang-Jin Choi and Hyeong-Seok Ko.
\newblock Online motion retargetting.
\newblock {\em The Journal of Visualization and Computer Animation},
  11(5):223--235, 2000.

\bibitem{feng2012automating}
Andrew Feng, Yazhou Huang, Yuyu Xu, and Ari Shapiro.
\newblock Automating the transfer of a generic set of behaviors onto a virtual
  character.
\newblock In {\em International Conference on Motion in Games}, pages 134--145.
  Springer, 2012.

\bibitem{gleicher1998retargetting}
Michael Gleicher.
\newblock Retargetting motion to new characters.
\newblock In {\em Proceedings of the 25th annual conference on Computer
  graphics and interactive techniques}, pages 33--42, 1998.

\bibitem{gomes2021shape}
Thiago~L Gomes, Renato Martins, Jo{\~a}o Ferreira, Rafael Azevedo, Guilherme
  Torres, and Erickson~R Nascimento.
\newblock A shape-aware retargeting approach to transfer human motion and
  appearance in monocular videos.
\newblock {\em International Journal of Computer Vision}, 129(7):2057--2075,
  2021.

\bibitem{goodfellow2020generative}
Ian Goodfellow, Jean Pouget-Abadie, Mehdi Mirza, Bing Xu, David Warde-Farley,
  Sherjil Ozair, Aaron Courville, and Yoshua Bengio.
\newblock Generative adversarial networks.
\newblock {\em Communications of the ACM}, 63(11):139--144, 2020.

\bibitem{jang2018variational}
Hanyoung Jang, Byungjun Kwon, Moonwon Yu, Seong~Uk Kim, and Jongmin Kim.
\newblock A variational u-net for motion retargeting.
\newblock In {\em SIGGRAPH Asia 2018 Posters}, pages 1--2. 2018.

\bibitem{jin2018aura}
Taeil Jin, Meekyoung Kim, and Sung-Hee Lee.
\newblock Aura mesh: Motion retargeting to preserve the spatial relationships
  between skinned characters.
\newblock In {\em Computer Graphics Forum}, volume~37, pages 311--320. Wiley
  Online Library, 2018.

\bibitem{kingma2015adam}
Diederik~P Kingma and Jimmy Ba.
\newblock Adam: A method for stochastic optimization.
\newblock In {\em The International Conference on Learning Representations},
  2015.

\bibitem{kulkarni2015deep}
Tejas~D Kulkarni, William~F Whitney, Pushmeet Kohli, and Josh Tenenbaum.
\newblock Deep convolutional inverse graphics network.
\newblock {\em Advances in neural information processing systems}, 28, 2015.

\bibitem{lee1999hierarchical}
Jehee Lee and Sung~Yong Shin.
\newblock A hierarchical approach to interactive motion editing for human-like
  figures.
\newblock In {\em Proceedings of the 26th annual conference on Computer
  graphics and interactive techniques}, pages 39--48, 1999.

\bibitem{li2022iterative}
Shujie Li, Lei Wang, Wei Jia, Yang Zhao, and Liping Zheng.
\newblock An iterative solution for improving the generalization ability of
  unsupervised skeleton motion retargeting.
\newblock {\em Computers \& Graphics}, 104:129--139, 2022.

\bibitem{lim2019pmnet}
Jongin Lim, Hyung~Jin Chang, and Jin~Young Choi.
\newblock Pmnet: learning of disentangled pose and movement for unsupervised
  motion retargeting.
\newblock In {\em 30th British Machine Vision Conference}. British Machine
  Vision Association, BMVA, 2019.

\bibitem{loper2015smpl}
Matthew Loper, Naureen Mahmood, Javier Romero, Gerard Pons-Moll, and Michael~J
  Black.
\newblock Smpl: A skinned multi-person linear model.
\newblock {\em ACM transactions on graphics}, 34(6):1--16, 2015.

\bibitem{paszke2019pytorch}
Adam Paszke, Sam Gross, Francisco Massa, Adam Lerer, James Bradbury, Gregory
  Chanan, Trevor Killeen, Zeming Lin, Natalia Gimelshein, Luca Antiga, et~al.
\newblock Pytorch: An imperative style, high-performance deep learning library.
\newblock {\em Advances in neural information processing systems}, 32, 2019.

\bibitem{peng2021animatable}
Sida Peng, Junting Dong, Qianqian Wang, Shangzhan Zhang, Qing Shuai, Xiaowei
  Zhou, and Hujun Bao.
\newblock Animatable neural radiance fields for modeling dynamic human bodies.
\newblock In {\em Proceedings of the IEEE/CVF International Conference on
  Computer Vision}, pages 14314--14323, 2021.

\bibitem{rempe2020contact}
Davis Rempe, Leonidas~J Guibas, Aaron Hertzmann, Bryan Russell, Ruben Villegas,
  and Jimei Yang.
\newblock Contact and human dynamics from monocular video.
\newblock In {\em European conference on computer vision}, pages 71--87.
  Springer, 2020.

\bibitem{ronneberger2015u}
Olaf Ronneberger, Philipp Fischer, and Thomas Brox.
\newblock U-net: Convolutional networks for biomedical image segmentation.
\newblock In {\em International Conference on Medical image computing and
  computer-assisted intervention}, pages 234--241. Springer, 2015.

\bibitem{tak2005physically}
Seyoon Tak and Hyeong-Seok Ko.
\newblock A physically-based motion retargeting filter.
\newblock {\em ACM Transactions on Graphics}, 24(1):98--117, 2005.

\bibitem{vaswani2017attention}
Ashish Vaswani, Noam Shazeer, Niki Parmar, Jakob Uszkoreit, Llion Jones,
  Aidan~N Gomez, {\L}ukasz Kaiser, and Illia Polosukhin.
\newblock Attention is all you need.
\newblock {\em Advances in neural information processing systems}, 30, 2017.

\bibitem{villegas2021contact}
Ruben Villegas, Duygu Ceylan, Aaron Hertzmann, Jimei Yang, and Jun Saito.
\newblock Contact-aware retargeting of skinned motion.
\newblock In {\em Proceedings of the IEEE/CVF International Conference on
  Computer Vision}, pages 9720--9729, 2021.

\bibitem{villegas2018neural}
Ruben Villegas, Jimei Yang, Duygu Ceylan, and Honglak Lee.
\newblock Neural kinematic networks for unsupervised motion retargetting.
\newblock In {\em Proceedings of the IEEE conference on computer vision and
  pattern recognition}, pages 8639--8648, 2018.

\bibitem{zhou2019continuity}
Yi Zhou, Connelly Barnes, Jingwan Lu, Jimei Yang, and Hao Li.
\newblock On the continuity of rotation representations in neural networks.
\newblock In {\em Proceedings of the IEEE/CVF Conference on Computer Vision and
  Pattern Recognition}, pages 5745--5753, 2019.

\end{thebibliography}

}

\end{document}